\documentclass[review]{elsarticle}

\usepackage{longtable}
\usepackage{subfigure}
\usepackage{setspace}
\usepackage{geometry} 
\usepackage{epstopdf}
\usepackage{amsmath}
\usepackage[labelsep=period]{caption}  
\usepackage{hyperref}
\usepackage[british]{babel} 
\usepackage[hang]{footmisc}

\usepackage{soul}
\usepackage{color}
\usepackage{url}

\usepackage[normalem]{ulem}
\usepackage{color}

\usepackage{enumitem}

\journal{Journal of Results in Engineering}

\begin{document}

\begin{frontmatter}

\title{Remote sensing and AI for building climate adaptation applications}





\author{Beril Sirmacek$^{1*}$}{\corref{mycorrespondingauthor}}
\author{Ricardo Vinuesa$^{2*}$}{\corref{mycorrespondingauthor}}
\cortext[mycorrespondingauthor]{Corresponding authors}

\address{$^1$ create4D, Overijssel, The Netherlands (e-mail: bsirmacek@gmail.com) \\
$^2$ FLOW, Engineering Mechanics, KTH Royal Institute of Technology, Stockholm, Sweden (e-mail: rvinuesa@mech.kth.se)
}

\begin{abstract}
Urban areas are not only one of the biggest contributors to climate change, but also they are one of the most vulnerable areas with high populations who would together experience the negative impacts.  In this paper, we address some of the opportunities brought by satellite remote sensing imaging and artificial intelligence (AI) in order to measure climate adaptation of cities automatically. We propose a framework combining AI and simulation which may be useful for extracting indicators from remote-sensing images and may help with predictive estimation of future states of these climate-adaptation-related indicators.  When such models become more robust and used in real-life applications, they may help decision makers and early responders to choose the best actions to sustain the well-being of society, natural resources and biodiversity. We underline that this is an open field and an on-going area of research for many scientists, therefore we offer an in-depth discussion on the challenges and limitations of data-driven methods and the predictive estimation models in general.
\end{abstract}


\end{frontmatter}


\section{Introduction}\label{Introduction}
 
Climate change is becoming a bigger threat for the well-being of all living beings on the planet day by day~\cite{mcpherson}. High ecological stress and the heat islands created by urban areas create a huge impact not only in the urban areas themselves but also in the surrounding rural areas because of the urban-heat-island (UHI) effect~\cite{bou_zeid}. While being one of the biggest contributors to climate change, cities are also one of the most vulnerable areas to the negative impacts. The United Nations predict that before 2050, $74\%$ of the European population and $68\%$ of the world population will be living in cities~\cite{UN2018}. Regarding these figures, if the current climate-emergency triggers of the urban areas are not well-identified and changed, a large number of the world population, and the biodiversity within and at the surrounding of the urban areas, will be in existential stress. 

For identification of the climate impact on the urban areas, many IT-infrastructured (also called `smart') cities, have been putting efforts and resources to collect a good amount of data which might be helpful to identify the climate-stress factors. Thus in many smart cities, citizens, government institutions, industry and scientists share data for the benefit of all; this relation is presented with the `Quintuple Helix model'~\cite{Carayannis2012}. Obviously, this leads to a great amount of data collection and the need for machine-learning (ML) or artificial-intelligence (AI) models that can use the data for extracting climate- and land-use-related indicators, which in turn could also be used for creating predictive models. We also need to acknowledge that not all cities are able to collect such big data from distributed sensors and from the mentioned data-sharing parties. Therefore, when possible, extracting indicators from satellite images would help to develop predictive models which could be used in any city globally.

If climate-change and human-activity-related indicators are extracted from remote sensing images, with power of AI methods, there would be possibilities for:
\begin{itemize}[noitemsep]
    \item Creating rapid maps of land use and environmental resources in large scales.
    \item Making predictions about future states of the extracted indicators.
    \item Simulating hypothetical scenarios for disaster prevention.
    \item Identifying abnormal situations (outlier identification).
    \item Explaining the impact of the indicators with eXplainable AI (XAI) methods.
    \item Identifying the relation between human activity, climate change and biodiversity-related changes.
    \item Real-time predictions and actions in an urban environment based on sparse measurements from within the city.
\end{itemize}

The list above could be extended further, however for now we will be limiting the discussion with these possible applications. Considering that, when such AI models are developed, they must immediately be used in real-life applications because of the climate emergency, in this article we would like to discuss also the following practical topics:
\begin{itemize}[noitemsep]
    \item Data collection.
    \item Feature extraction.
    \item Model selection.
    \item Generalization.
    \item Reproducibility.
    \item Maintainability.
    \item Simulation tools at different scales.
\end{itemize}

The focus here is on developing a sustainable smart city development perspective, keeping the sustainable development goals (SDGs), as well as the ethical \& responsible use of AI~\cite{vinuesa_nat,gupta_ai} as the ultimate judgement criteria for each of our suggestions. Because of their high scalability, low cost and reliable data-collection properties, we will discuss the possibilities of using satellite images as data source, as well as simulation data of multiple resolutions. In the rest of the paper, we will answer the following research questions:

\begin{enumerate}[noitemsep]
\item How can remote-sensing images and AI help to extract land-use and environment-related indicators to monitor climate adaptation of cities? 
\item How could fine-scale flow-simulation and climate-modelling approaches, as well as robust prediction models be achievable?
\item What are the limitations of remote-sensing images and AI models for adequate climate-adaptation monitoring and for performing accurate predictions?
\end{enumerate}

We assume that the reader acknowledges that the current climate emergency (anthropogenic climate change) is mainly caused by  humans  (i.e. daily consumption and commuting choices, industrial  activities, etc.). Therefore, one of our main ambitions with this paper is to suggest observation systems which can highlight the relation between human land use and environmental changes. In this way, we hope that the climate- change- related  emergency  could  be  understood and the negative impacts could be decreased by changing activities which contribute to this environmental stress. Nevertheless, expectations from AI models about solving climate-related problems should be realistic. In the discussion section we discuss both strengths and weaknesses of AI models when it comes to implementing fully-automated applications with real-life data.

\section{Background}\label{sec:background}

\subsection{Climate models, remote sensing and the impact of AI models}

As the Intergovernmental Panel on Climate Change (IPCC) has been warning in its reports, climate change is likely to bring devastating consequences to the health of humans and animals, to social living and to environmental resources~\cite{ipcc2018}. The accelerated speed of climate change could be slowed down with significant reduction of greenhouse gas emissions and the heat-island impacts created by the urban areas \cite{gge2018,bou_zeid}. Levels of carbon dioxide in the atmosphere have remained around a narrow range over the last million years. In the last hundred years, they have risen from 280 ppm (parts per million) to 400 ppm. These changes in CO$_2$ concentration closely match with the human-made developments within urban areas~\cite{Berkley2014}.

In this context, it is important to note that around 75\% of the population currently lives in cities in the European Union (EU) and it is expected that, by 2030, 60\% of the population will be urban dwellers worldwide~\cite{unga2}. Cities are responsible for a significant fraction of the total CO$_2$ emissions in the world (from 60 to 80\%), and they are expected to play a prominent role towards the achievement of the climate-change targets from the Paris Agreement~\cite{paris3}. When it comes to air pollution, the European Environment Agency (EEA) reports that around 90\% of the urban population in the EU was exposed to levels exceeding the ones recommended by the World Health Organization (WHO) between 2014 and 2016, and those levels of air pollution are responsible for 800,000 premature deaths~\cite{deaths800k}. On the other hand, the UHI phenomenon was related to 70,000 deaths as a consequence of the heat wave in Europe during the summer of 2003~\cite{heat6}. Although the EU has introduced the use of predictive models to support pollutant-concentration measurements~\cite{air7}, currently available techniques are unable to provide the spatial and temporal accuracy required to reproduce the pollutant dispersion patterns within urban environments~\cite{carpentieri8}. Similarly, more robust non-intrusive-sensing methods need to be developed in order to accurately predict the heat fluxes relevant to the UHI phenomenon~\cite{weng9}. Therefore, there is a pressing need for improved prediction and assessment methods to tackle these challenges and enable urban sustainability in the near future. This has important implications on the Sustainable Development Goals (SDGs) 11 (on sustainable cities and communities) and 13 (on climate change) from the United Nations (UN) 2030 Agenda~\cite{unga2,vinuesa_nat}. Significant progress can be made with adequate simulation tools to properly understand and predict the complex flow in urban areas~\cite{torres,stuck}.

It is also important to note that Satterthwaite stated: \textit{``Do not blame cities"}~\cite{Satterthwaite2008} for being the biggest stressors of climate change because oil refineries, deforestation activities, animal agriculture, heavily chemical and fossil-fuel using agriculture, commercial transportation etc. take place outside the cities. However, we should not ignore the fact that those activities within the rural areas happen for the purpose of bringing food, fossil fuel and other needs of the highly-industrialized and densely-populated areas within the cities \cite{Kennedy2007}. Thus, even though such big climate-stressing activities take place outside the cities, they are associated with the lifestyle and demand within the cities. Therefore, we believe that it is extremely valuable to focus on cities in climate-related studies.

A proper assessment of the climate-change stressors of the urban areas requires observation of the conditions of these areas in multiple aspects. Soil/water surface temperatures, atmospheric events, vegetation/ice cover are just a few of them. Many climate models can predict future states of the climate for a region using these parameters~\cite{randall2007}. Two major models are frequently used in this field are the Earth System Models (ESMs) and the Global Climate Models (GCMs)~\cite{climatemodels}. ESMs include all the features of GCMs and also simulate the carbon cycle and other chemical and biological cycles that are important for determining the future concentrations of greenhouse gases in the atmosphere. ESM models simulate the environmental indicators in large computational domains and they enable to make predictions for large regions in an acceptable performance. However, when it comes to performing prediction for a finer scale (e.g. a city-size area), the results become less accurate. This is the reason why it is common to see news articles reporting that the ice sheet of a certain area is melting faster than what scientists have expected~\cite{arctic2020}. This effect was studied by Schenk et al.~\cite{schenk}, who performed finer-resolution ESM simulations, and identified very relevant flow-blockage effects~\cite{schenk_Vinuesa} absent in the coarse cases, and very relevant to the global climate behavior. Bader et al.~\cite{bader2008} also discussed this issue, reporting that the global models (or large-grid models) do not provide very accurate results when they are used in a smaller region. If the models are going to be used in a city-size focus region, the model parameters need to be adapted regarding the human-made developments and activities in that area. They added that, even then these fine-tuned models will exhibit adequate performance for the focus regions only for a short time frame, because they will be missing the relation to the global changes. Arctic ice melt and Amazon rain forest damages would eventually make an impact in a far away region. However, the fine-tuned local models would not account for this crucial connection to the global events. Last but not least, Bader et al.~\cite{bader2008} addressed the extremely high time consumption and costs of fine-tuning climate models for every small area globally. 

To make models more successful for local applications, Asch et al.~\cite{Asch2016} suggested that the most suitable way to make predictions would be simulating many different scenarios about the indicators of a specific region. Kitchin et al.~\cite{KitchinStehle} addressed various different type of data that should be collected in smart cities for better understanding of the causes and effects between the activities of large populations and the environment. They also discussed challenges of collecting such specific data variety within urban areas. Many climate scientist agree that representative urban-region indicators for tracking human-made developments and environmental changes should be added into climate models. These models could also be helpful for understanding whether a city is developing within environmental limits or not. For this purpose, Caird et al.~\cite{Caird2019} suggested that ISO (International Organization for Standardization) smart-city indicators (ISO 37122:2019 Sustainable cities and communities, Indicators for smart cities) could be useful to identify what measurements should be collected in order to track sustainable development of cities.

Earlier in the literature, some researchers have focused on collecting internet-of-things (IoT) data from cities in order to create big data to measure sustainable development~\cite{Giang,Su}. Usage of remote-sensing data for sustainable smart-city-development measurement has been considered by some scientists as well. Bonafoni et al. \cite{BONAFONI2017211} showed that it is possible to measure surface heat stress of cities using satellite sensors in order to model and track the changes of the urban heat island effect. Milojevic-Dupont and Creutzig \cite{MILOJEVICDUPONT2021102526} have suggested that remote-sensing images could be used for measuring air quality, biomass, carbon, soil/water surface temperatures, built up area growth and water quality/levels. It is still not clearly identified how to validate and generalize these methods and what universal measures to use in order to define the concepts of ``climate adaptation'' and ``sustainability'' in the context of the smart cities. 

G\"uemes et al.~\cite{guemes2021coarse} have shown that high-fidelity turbulent-flow simulations can be used together with AI models when coarse measurements are collected. This study increases trust in future possibilities of high-fidelity flow and climate modeling using remote-sensing images and AI models. Note that the complexity of the turbulent flows present in these urban environments, and the sheer amount of produced data, require the use of deep-learning models to be able to obtain accurate and robust predictions. It has been shown that nonlinear methods, including approaches based on deep learning, outperform classical ones when modeling complex fluid-flow cases ~\cite{guastoni_et_al,kenzo}. Some additional techniques in this context include fine-tuned convolutional and recurrent neural networks combined with transfer learning~\cite{srinivasan_et_al,guastoni_et_al}. The potential of high-fidelity simulations for non-intrusive sensing in urban areas is discussed below.

In this paper, we propose methods to extract urban-development and environmental indicators from both remote-sensing images and non-intrusive real-time data. We argue that both are very helpful for assessing climate adaptation of cities. Besides, we suggest AI frameworks which may be helpful for creating fine-detail flow and climate models for smaller-interest regions and to predict their future climate status. As a complement to this, we discuss that satellite images not only provide large-scale (planet-wide) observation possibilities but also they offer the most sustainable observation solution without need of in-situ sensor and embedded system installation.

\subsection{Smart-city indicators}

\begin{figure*}[ht!]
\begin{center}
		\includegraphics[width=.9\columnwidth]{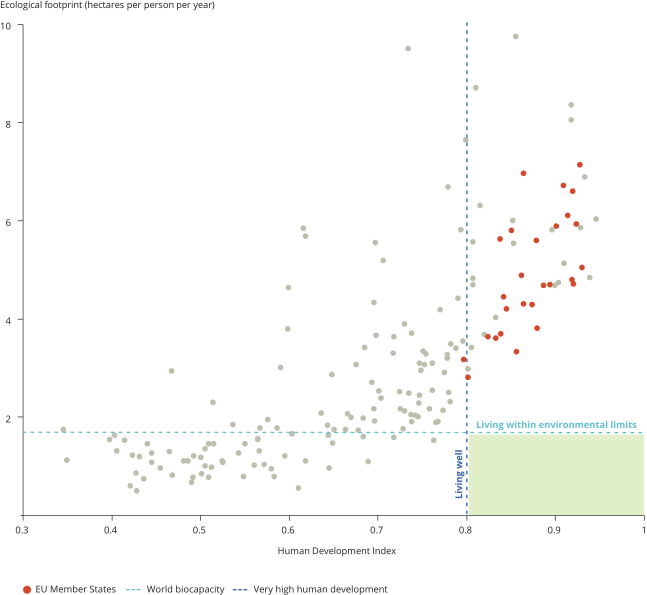}
	\caption{Correlation between Ecological Footprint and Human Development Index (HDI). The HDI values above 0.8 are defined as ``Living well'', whereas Ecological footprint values below 1.7 are defined as ``Living within environmental limits''. The value of 1.7 refers to the average biocapacity per person globally in 2014. Red and gray data points represent EU-member and non-EU-member countries respectively. Figure adapted from: European Environment Agency \url{https://www.eea.europa.eu/data-and-maps/figures/correlation-between-ecological-footprint-and}.}
\label{fig:eea_example}
\end{center}
\end{figure*}

The European Environment Agency (EEA) has published indicators which could be used for observing the sustainability status of the smart cities \cite{eea2019}. As in the example given in Figure~\ref{fig:eea_example}, the EEA has suggested that correlation between the ecological footprint and human development indices (HDI) could be used for measuring the sustainability levels and climate adaptation of countries. Detailed mathematical processes for calculation of the ecological footprint and HDI indices have been introduced in the EEA technical note document~\cite{eea2019indicators}.

In this figure, the Ecological Footprint is the metric that shows how much natural resources are being used (or damaged) to support people and their economical/industrial activities in a region. Ecological-footprint calculation with IoT data is explained within the documentation in detail~\cite{ecoapi}. The horizontal axis, Human Development Index (HDI), is a summary measure of average achievement in key dimensions of human development: a long and healthy life, being knowledgeable and having a decent standard of living. The HDI is the geometric mean of normalized indices for these three dimensions~\cite{eea2019indicators}. When these two indicators (one focused on the human-development activities and another one on the natural wellness) are available at a finer scale, a diagram like the one in Figure~\ref{fig:eea_example} can help to track the climate adaptation of a city assuming that climate adaptation requires a good balance between these two indicators (the green area in the figure). Furthermore, these indicators could contribute to developing fine-tuned flow and climate models which could be used for prediction of the future climate status of the city (see \S\ref{sec:howAIcan}).

Unfortunately, EEA indicators cannot be used in an automated framework for the time being, because of the complexity of the parameters which are used to calculate the Ecological Footprint and HDI indicators. Obviously, a parameter like human health or human long life cannot be extracted from satellite images or flow simulations. It could be possible to automate a model with the model development process if these parameters are provided through an application programming interface (API) from each city, a possibility that is not currently realistic since different cities have different budgets to construct such a database and API. For observing such ecological and human-development indicators, the ISO proposed a set of indicators which could be standard for each city~\cite{ISO}. However, the ISO indicators also rely on availability of a database provided by the city. 
In order to discuss automation possibilities of the ecological and human-development observation process, being inspired by the EEA and ISO indicators, in Table~\ref{smartcity_indicators} we list the city indicators which could directly or indirectly be observed by satellite images. When it is possible to measure indicators directly as it was described within the documentation, the indicator is labelled as `direct'. When it is not possible to measure the indicator directly as it is described, but it is possible to extract highly-correlated measurements, the indicator is labelled as `indirect'. For instance, if the city tracks the number of people who have completed a high education and provides this number with an API, that could be a direct IoT measurement. As another example, economical indicators of a city cannot be directly inferred from a satellite image, but it could be possible estimate them based on building and road construction. In other words, direct refers to the methods which directly produce the indicator value, whereas indirect refers to the methods which rely on other properties within the data in order to estimate the indicator value. Assuming that all the indicators could be provided as direct measurements via a city database if there was such smart-city infrastructure available, remote sensing-data enables fewer measurements and more indirect indicators than IoT measurements. However, remote-sensing data would have advantages over IoT data such as having a lower privacy concern, providing large-area observations and providing the opportunity to repeat the measurements consistently as the satellite observations are done automatically. Satellite-image based observation also provides advantage for the less developed cities because of not demanding any sensor installation and back-end development for creating a database and API.
 
\begin{longtable}{|p{0.3cm}|p{2.5cm}|p{1.6cm}|p{6.4cm}|p{1.8cm}|p{1.2cm}|}
		 \hline
			No & Smart-city indicator & Remote sensing & Method & References & SDG \\\hline
			1 & Economy & Indirect & Residence-building shape and size, gardens (swimming pools, golf courts, etc.) or slum-looking texture can give indication about the economic status of a region. & \cite{Kuffer2016} & 8 \\\hline
			2 & Education & Indirect & Population density, greenness and development of the city (well-organized buildings and large roads) might indicate higher ratios of well educated community.  & \cite{weng2007} & 4 \\\hline
			3 & Energy & Indirect & Building size and their estimated human capacity can be used to predict the amount of energy consumption. & \cite{LIU2021127} & 7\\\hline
			4 & Environment and climate change & Direct & Satellite images can show the deforestation, ice and vegetation cover changes. & \cite{Dutrieux2012} & 15\\\hline
			5 & Poverty and health & Indirect & Satellite images can help to predict availability of food resources and their variety. This might indicate the type of malnutrition and potential illnesses in the area. Besides, the air-quality observation might shed light into the respiratory diseases.  & \cite{Bondi2020MappingFP}, \cite{cesar2019}, \cite{jean_science} & 1,3 \\\hline
			6 & Recreation & Indirect & Convolutional neural networks (CNNs) can be trained to recognize parks and other recreational buildings/areas from satellite images. Nevertheless, many recreation activities are held indoors, therefore we labelled this indicator as an indirect one. & \cite{hu2016}& 3\\\hline
			7 & Safety & Indirect & Incident locations can be labelled on a map and an ML algorithm can be trained to recognize potential incident locations by looking at road, building, green area indicating features. & \cite{Najjar}& 16\\\hline
			8 & Telecommunication & Direct & Satellite images give possibility to identify towers which enable various communication channels. & \cite{nico2020} & 9\\\hline
			9 & Transportation & Direct  & Computer vision, ML and AI methods can be used to extract road network, seaports and airports from satellite images. Besides, when Synthetic-aperture radar (SAR) images are available, they can indicate the motion direction and speed of the transport vehicles. & \cite{unsalan2012}, \cite{chen2018}, \cite{hoppe2016} & 9\\\hline
			10 & Agriculture and food security & Direct & Type of yield and their growth status can be recognized by satellite-image processing and machine-learning models can help to estimate how much yield will be available on which date. & \cite{Sayago2018}& 2\\\hline
			11 & Waste & Direct & Satellite images can provide visual information to detect the amount of waste in open-collection areas. However, the collections in the close areas and the separability of the waste cannot be known. Nevertheless, the open area waste could provide some indication about the amount of waste generated. & \cite{Vambol2019}& 12\\\hline
			12 & Water & Direct & Water-quality parameters: suspended sediments (turbidity), chlorophyll and temperature can be identified by satellite images. & \cite{ross2019}, \cite{Ritchie2003} & 6, 14\\\hline
	\caption{Smart-city indicators to measure climate adaptation and some potential satellite-image-based methods to extract those indicators automatically.}
\label{smartcity_indicators}
\end{longtable}

\section{Methods}\label{sec:howAIcan}

In order to answer the research questions, we propose designing the following four modules:

\begin{enumerate}[noitemsep]
\item Extract land-use and ecological indicators from satellite images.
\item Climate-adaptation observation.
\item Train predictive models.
\item High-fidelity finer-scale flow and climate simulations.
\end{enumerate}

\begin{figure}[ht!]
\begin{center}
	\includegraphics[width=.7\columnwidth]{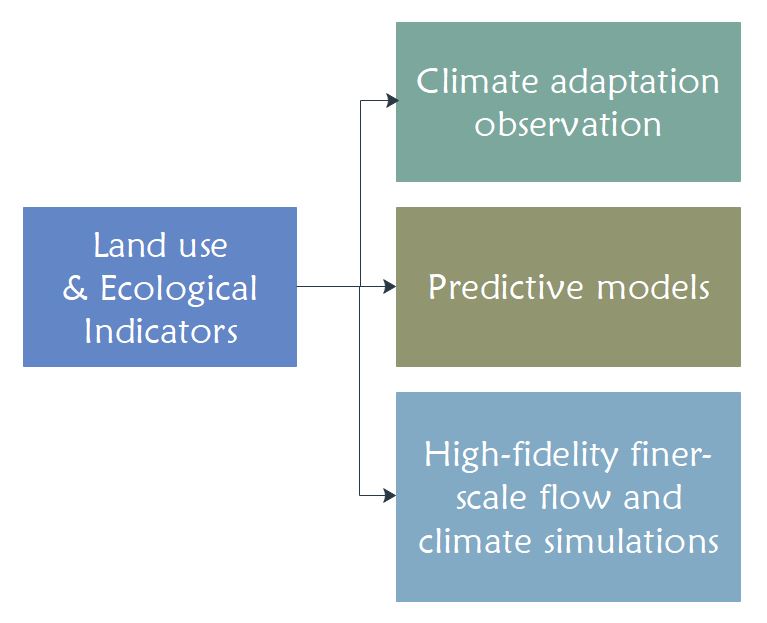}
	\caption{Modules proposed in this study towards answering the research questions.}
\label{fig:modules}
\end{center}
\end{figure}

We can summarize the relation between these modules as in Figure~\ref{fig:modules}. Next, we discuss development of each module. 

\subsection{Extract land-use and ecological indicators from satellite images}

For building a successful model, it is important to follow the ecological footprint index and HDI formulas suggested by EEA. In Table~\ref{smartcity_indicators} we discussed availability of the indicators which serve as the parameters of these formulas when the satellite images are used as a resource. Even though it is promising to acknowledge that satellite images could provide either direct or indirect measurements for most of these parameters, significant research is needed to understand how to calibrate and normalize the satellite-image-based indicators before providing their values to these formulas. A thorough assessment of this will be devoted to future work, and in the current contribution we rely on simpler formulas to replace the ecological-footprint index and HDI. 

For the sake of designing the first practical model, we chose a simple indicator which may roughly represent HDI. This is the land-development index which was proposed in a previous research work~\cite{SIRMACEK20101155}. Again for the sake of simplicity, we use the urban green index~\cite{Gupta2012} instead of the ecological footprint index, since the latter requires many more parameters to calibrate, normalize and use in the calculations. It is clear that these two parameters are significantly simplified indices to replace the EEA indices, nevertheless they may help to develop the first automated models before more research is conducted to discover better representations of the EEA indices using satellite images. 


\subsection{Climate-adaptation observation}
 
In order to assess the extent to which the city is aligned with the climate goals, we propose to use a diagram similar to that shown in Figure~\ref{fig:eea_example}. For our simplified-index-based setup, the vertical axis should be represented with the inverse of the urban green index (since in the original plot, it represents the footprint, not the greenness) and the horizontal axis should be represented with the land-development index. The largest challenge would be to calibrate and normalize these two indices. To this end, we propose to calculate the urban green index and the land-development indices of a few countries (not cities), and use the Figure~\ref{fig:eea_example} table values for calibrating these satellite-imaging-based indicators. Once the calibration parameters are known, these can be used to represent the city indicators in a similar figure. The latest SDG ranking report~\cite{sdgranking} could also be used for judging the trustworthiness of the satellite-imagery-based new ranking results. It would be another future research focus of ours to find good threshold values to represent the boundaries of this green area in the satellite-based measurement plot. For now, the threshold values could be assumed as $20\%$ of each axis, similar to the EEA-index-based plot. Since after the normalization process, the vertical and the horizontal axis values will be between $[0,1]$, the horizontal-axis threshold value would be $0.8$ and the vertical-axis threshold value would be $0.2$. Nevertheless, these satellite-based observations must be repeated on many cities, in order to determine whether these threshold values are still representative of the climate-adaptation area or not. We suggest that when the satellite-based indicators are obtained, instead of making direct comparison with the EEA-index-based plot, the satellite-based indicators should be used in a new plot. In such a new plot, climate adaptation developments of two different regions, or climate adaptation development of the same region in different times, could be judged with respect to each other. Currently, none of the countries in the world appears in the green area, which is assumed to represent a good status for climate adaptation. However, when finer-scale calculations are conducted, it might be possible to observe some cities which might lay within the green area. We hope that such cities could also be used as an example to understand how to keep `living well' and `living within environmental limits' in a good balance. Automatic observation provides opportunities to easily track the changing status of the cities in the figure.
 
\subsection{Train predictive models}

In forecasting applications AI uses a time series of historical data to learn parameters of a good fitting model representing the trend of this database in time. After the model is trained the same model is used to find the data points in any future time. Thus, when the urban green index and the land-development index used in the climate-adaptation observation module are collected for a time period (for instance every month for a city), it would be possible to use ML or deep-learning models to create a prediction model to estimate future values of the urban green index and the land development. 

Before choosing the right prediction model, it is important to analyze the existing training data in order to assess the data behavior. If the trends could be represented with a polynomial function, it would be so much easier to develop a regression model with ML algorithms. This process would also require fewer training examples. On the other hand, if the trends of different cities do not look like showing a similar increase or decrease and if they look too complex to be presented by a polynomial function, it would be a better idea to use a deep-learning method for predictive modeling. For such a case, training a Long short-term memory (LSTM)~\cite{lstm1997} network would help to achieve more robust results ~\cite{srinivasan_et_al}. LSTMs are successful in remembering what changes happen after significant events in the past and they are less likely to be affected by the noise in the data, which may be produced by a short-term and irrelevant event.

\subsection{High-fidelity finer-scale flow and climate simulations}
\begin{center}
\textit{``All models are wrong, but some are useful. --- George Box"}
\end{center}

In the background section we discussed the strengths and weaknesses of the existing climate models which have been used for decades. To summarize:

\begin{itemize}
    \item The existing climate models make reliable predictions~\cite{carbonbrief}.
    \item The existing climate models lack great amount of detail, having a coarse spatial resolution with a grid-cell size on the order of $2.5^{\circ}\times 2.5^{\circ}$ (approximately $275 \times 275$ km$^2$), which is far too coarse for identifying the impact of human activities, land use and the biodiversity threats~\cite{Flint2012}.
    \item AI-based methods, as well as high-fidelity flow simulations, may help to develop finer-scale models.
\end{itemize}

Here we first propose a framework for high-fidelity, fine-scale simulations ~\cite{vinuesa_review} at the city level, where the data will be used for non-intrusive real-time applications. Then, a wide-and-deep-learning approach is proposed to combine the local and global indicators in the predictions.

\subsubsection{Non-intrusive sensing in fine-scale urban data}

Here we propose to perform high-fidelity simulations by means of high-order numerical methods, which are adequate to simulate turbulent urban flows because of the wide range of scales present in turbulence. This type of simulation can be performed very efficiently by means of the spectral-element code Nek5000 \cite{fischer_et_al}. The spectral-element method (SEM) allows to combine the geometrical flexibility of finite elements with the accuracy of spectral methods, and it is possible to obtain very high levels of accuracy in urban-flow simulations~\cite{stuck}. The high-quality numerical data can be used to train deep-learning models for non-intrusive sensing, providing real-time flow fields based on sparse measurements. Note that the work in Ref.~\cite{stuck}  corresponds to well-resolved large-eddy simulations (LESs), which constitute high-fidelity simulations of turbulent flows. This case exhibits moderate geometrical complexity, but more realistic geometries can be considered with simulations of lower fidelity, such as Reynolds-averaged Navier--Stokes (RANS) approaches~\cite{clara}. An effective approach to benefit from the accuracy of high fidelity and the capacity to model realistic geometries of low fidelity is the multi-fidelity framework~ \cite{saleh}. We can reconstruct the flow field in two ways: in the first one ({\rm Reconstruction 1}) we decompose the flow by means of proper orthogonal decomposition (POD)~\cite{ref:pod}, which for $N+1$ three-dimensional snapshots reads as follows:
\begin{equation}
  \pmb{u}(\pmb{x},t)=\overline{\pmb{u}}(\pmb{x})+\sum_{i=1}^{N} a_i^u(t) \pmb{\phi}_i^u(\pmb{x}).  
\end{equation}
Here $\pmb{u}$ is the flow velocity vector, $\pmb{x}$ are the spatial coordinates and $t$ is time. Furthermore, $\overline{\pmb{u}}$ denotes the mean flow, $a_i^u(t)$ are the temporal coefficients and $\pmb{\phi}_i^u$ are the POD modes. In {\rm Reconstruction 1} we focus on predicting the temporal coefficients $a_i^u(t)$, given the calculated POD modes $\pmb{\phi}_i^u$, to construct the instantaneous behavior of the flow velocity $\pmb{u}$. Note that although here we illustrate the methods with the velocity field, the same approaches can be employed for reconstructing the temperature field and the concentration of pollutants (treated as passive scalars). The second alternative ({\rm Reconstruction 2}) aims to reconstruct the flow on a horizontal or vertical plane without performing any decomposition. 

\begin{figure*}[ht!]
\begin{center}
	\includegraphics[width=1.\columnwidth]{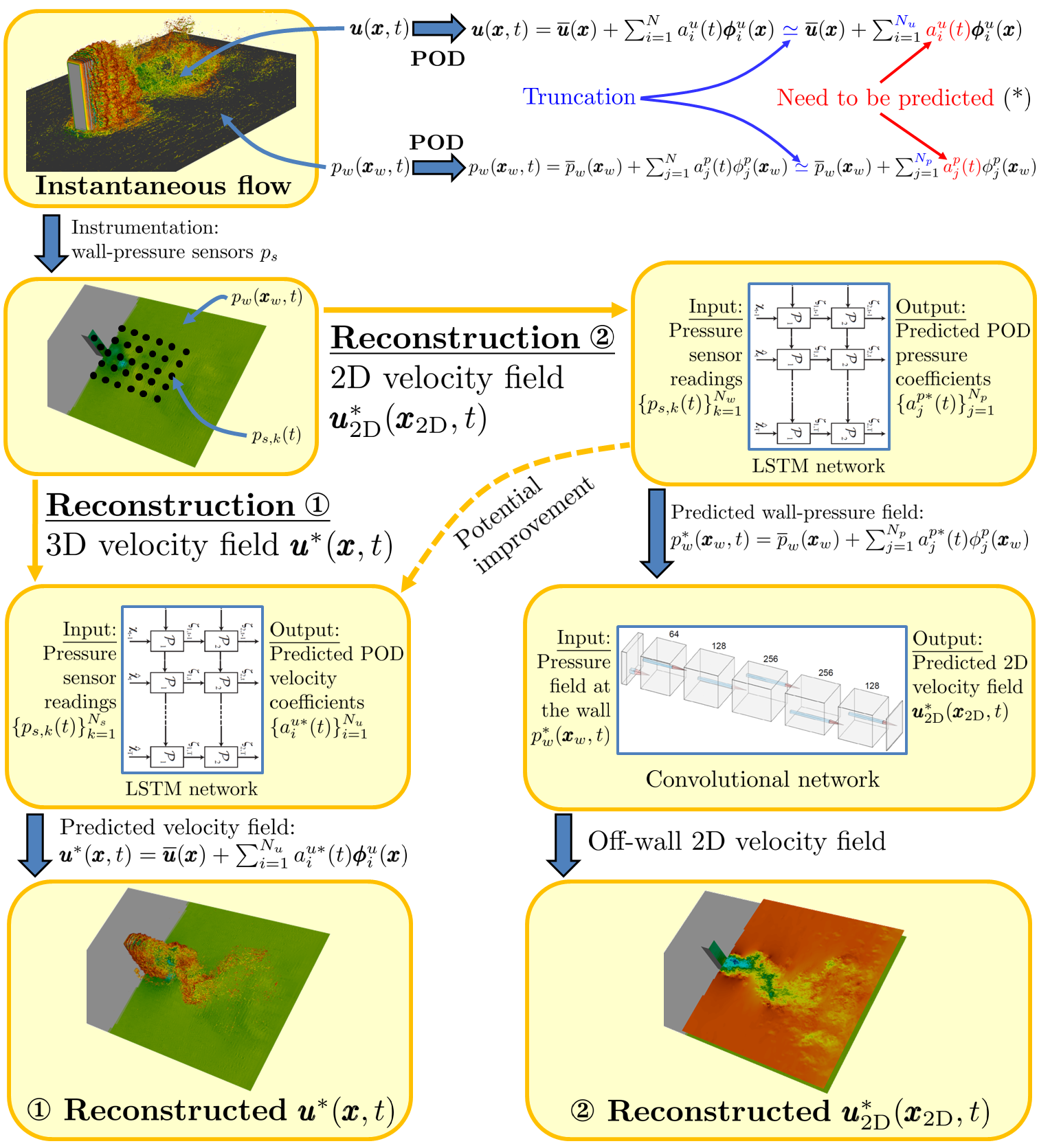}
	\caption{Schematic representation of the two proposed reconstruction methods, using high-fidelity simulation data of the wake from a wall-mounted square cylinder~\protect \cite{vinuesa_jot}. Here predicted quantities are denoted by $*$, and $p_w$ indicates wall pressure. This illustrative example can be extended to predictions of pollutant concentration or temperature distributions based on sparse measurements of pressure and/or temperature, and the framework can be applied to complex urban geometries. Note that $N_u$ and $N_p$ are the number of retained modes in the truncated POD expansions of the velocity and wall-pressure fields, respectively.}
\label{fig:reconstruction}
\end{center}
\end{figure*}

In Figure~\ref{fig:reconstruction} we illustrate the reconstruction process for the wake of a wall-mounted obstacle~\cite{vinuesa_jot}, where the temporal pressure measurements$\{ p_{s,k}(t) \}_{k=1}^{N_{s}}$ from $N_s$ sensors located at $\pmb{x}_s$ (on the street and obstacle surfaces) are used to predict the three-dimensional velocity field $\pmb{u}(\pmb{x},t)$. Note that this illustrative example can be extended to more general reconstructions, such as predictions based on sparse pressure and/or temperature measurements, and that this framework can be applied to more complex urban geometries. We use POD to perform the corresponding flow decompositions because, since it extracts orthogonal basis functions ranked by energy content, it requires fewer modes to reconstruct significant fractions of the energy content in the original flow than other approaches~\cite{Ref58,Ref59}. Alternative deep-learning approaches can also provide excellent reconstruction results from sparse measurements.~\cite{PEREZ2020109239,ABADIAHEREDIA2022115910} 

\subsubsection{Wide and deep learning}

\begin{figure*}[ht!]
\begin{center}
	\includegraphics[width=1.\columnwidth]{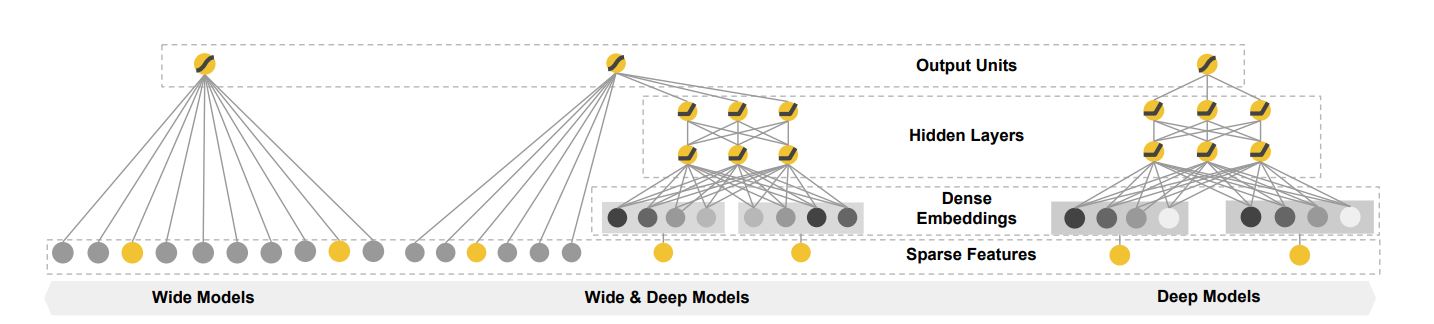}
	\includegraphics[width=.7\columnwidth]{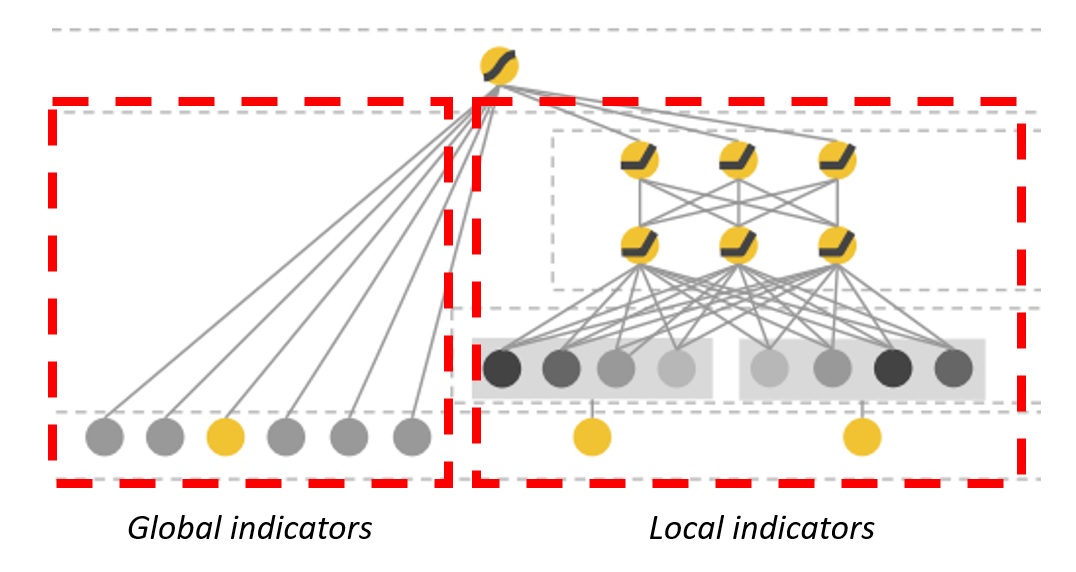}
	\caption{Schematic figure showing wide-and-deep-learning architecture and illustrating its use for building finer-scale flow and climate models. The deep neural network image is adapted from \protect\cite{cheng2016wide}.}
\label{fig:widedeep}
\end{center}
\end{figure*}

Cheng et al.~\cite{cheng2016wide} proposed a new AI model called `wide and deep learning', which significantly improved recommender systems. They provided a TensorFlow API to speed up research in this field ~\cite{tfwideapi}, a fact that allowed more researchers to test the successful practice of combining features of a deep neural network and the raw input together. The reason is that some input information might need to be processed in the deep neural network to find the most meaningful information within and some input information may be used unchanged. In our case of fine-scale flow and climate modeling, we believe that it is also beneficial to consider such a model as illustrated in Figure~\ref{fig:widedeep}. As future work, we believe that not only the urban green and the land-development indices, but all  indicators proposed in Table~\ref{smartcity_indicators} should be used as local indicators (deep model input). Furthermore, the existing climate-model parameters should be used as global indicators (wide model input). As an example, such a model could be trained for predicting the local temperatures or biodiversity variety. It could also be a part of the future work to investigate the importance of both local and global indicators using eXplainable-AI (XAI) techniques. In this context, some recent work has highlighted the importance of interpretable AI models towards achieving sustainability goals~\cite{vinuesa_interp}.

\section{Discussion}\label{sec:howAIcannot}

In the following subsections, we discuss research questions in light of the methods proposed in this study.

\subsection{How can remote-sensing images and AI help to extract land-use and environment-related indicators to monitor climate adaptation of cities? }

Remote-sensing images can help by monitoring the climate adaptation of the smart cities. AI allows predicting future states of the environmental and land development indicators of the cities. Besides, innovative AI models (like the use case of the wide-and-deep-learning method) can help with achieving finer-scale climate models to assess the climate indicators from the cities more efficiently.  

\begin{figure*}[h]
\begin{center}
		\includegraphics[width=0.9\columnwidth]{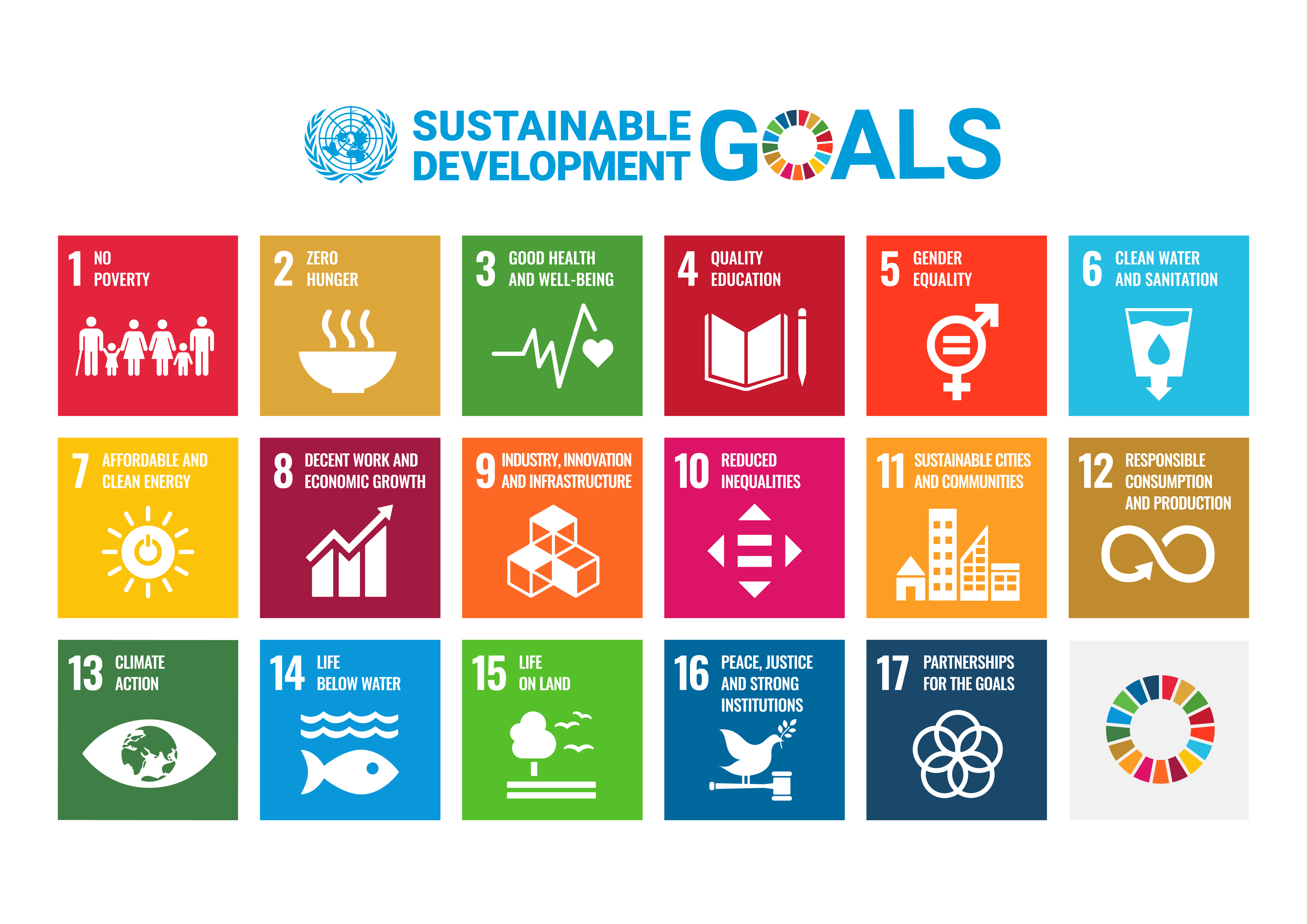}
		\includegraphics[width=0.6\columnwidth]{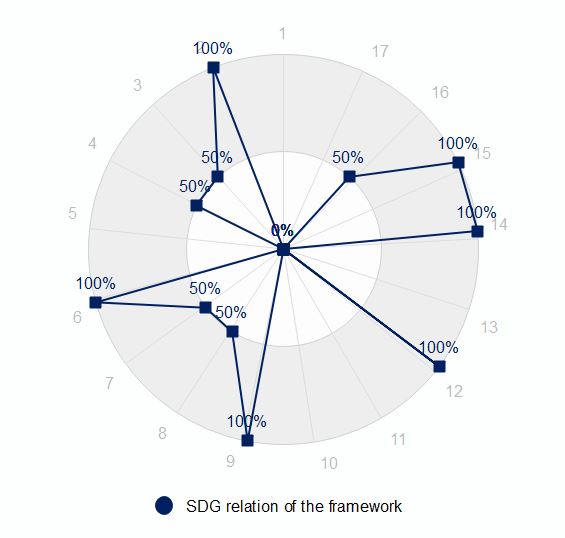}
	\caption{(Top) Summary of Sustainable Development Goals (SDGs) proposed by the United Nations. Panel adapted from: \protect\url{https://sdgs.un.org/goals}. (Bottom) Radar graph showing how remote sensing and AI-based methods can contribute to each SDG. Indirect impact is shown with $50\%$ and direct impact is shown with $100\%$ values on the graph. Additional information: \protect \cite{vinuesa_nat}.}
	\label{SDG}
\end{center}
\end{figure*}

When the modules in Figure~\ref{fig:modules} are developed, their outcomes can help with the SDGs listed in Table~\ref{smartcity_indicators}. Figure~\ref{SDG} shows the sustainable development goals which can be supported with satellite images and AI methods together. The radar chart in the figure illustrates how much help can be provided to each SDG. Herein, if the indicator extraction method is labelled as indirect in Table~\ref{smartcity_indicators}, the contribution to the SDG is represented as $50\%$, otherwise the contribution is represented as $100\%$. Additional references on this can be found in the work by \cite{vinuesa_nat}.

\subsection{How can innovative fine-scale flow and climate modeling, as well as prediction models, be achievable?}

As one of the potential solutions to achieve finer-scale flow and climate models, herein we propose an AI method known as wide and deep learning. However the performance of such models should be further investigated in future studies. The non-intrusive-sensing framework has the potential to provide high-quality real-time data to monitor air quality and thermal comfort in cities worldwide. 

\subsection{ What are the limitations of remote sensing images and AI models for adequate climate-adaptation monitoring and for conducting accurate predictions?}

Limitations of the remote-sensing and AI-based solutions for climate adaptation and prediction topics suffer from two main issues. One is the practical difficulties of building a fully automated framework with AI, and the second one is associated with the unknowns regarding climate change and climate-modeling science. Table~ \ref{tab:datawise_challenges_opportunities} summarizes the discussion on the challenges and limitations of the proposed methods. It also includes suggestions of potential solutions to further investigate these relevant topics.

\begin{longtable}{|p{2.5cm}|p{5cm}|p{6.5cm}|}
		\hline
			Earth data&Challenges&Solutions\\\hline
			 Training data& The measurements which happened earlier cannot be reproduced & If it was possible to save measurements, it would be possible to repeat the experiments when new models are developed. However, tracking the time variance of all indicators around the world would generate a very big data set which would cause other practical problems about storing data. 
			 \\\hline
			 Test and Evaluation data& The satellite sensors are improved every few years. They provide higher-quality and higher-resolution images. However, changing sensors would create a challenge for the models which were trained with lower resolution and lower quality data in the past. & Models possibly need to be re-trained to prevent deterioration. In this context, transfer learning~\cite{guastoni_et_al} may be a suitable path to tackle this issue. \\\hline
			 Model selection& Unfortunately, there is no formula which tells what AI/ML model to choose for each module. & The most common approach is to determine a few number of models which are assumed to be useful. This step, of course, depends on the intuition and experience of the developer. Afterwards, those candidate models could be compared in terms of their performance on training/testing data sets, their bias/variance and the number of the hyperparameters. When the response time is important, that could be taken into account as well. In environmental perspective, we believe that the developer must consider the trade of between the model performance and the hyperparameter numbers.\\\hline
			 Generalization& One model might not be suitable to represent any city around the world. & It might be also possible to segment the cities depending on their geographic conditions and train a separate model for each segment. \\\hline
			 Reproducibility& In practice an ML/AI framework should provide reproducible results. However, lack of reproducibility of the past test data might make this goal unachievable. & It would be possible to reproduce the results when the test data, model architecture and the model weights are stored properly. \\\hline
			 Maintainability& For many ML/AI frameworks, maintainability of the models is one of the biggest challenges. The most concerning maintainability issue is that models deteriorate in time (because input data changes with environmental conditions).  & Model deterioration might be overcome by re-training the models with new data in certain periods of time. Automatically updated satellite observations, like those from Copernicus services \cite{copernicus}, provide opportunities for re-training the models in certain frequency or when large environmental changes are noticed. \\\hline
	\caption{Practical challenges of implementing a remote-sensing and AI-based fully-automated framework for monitoring cities.}
\label{tab:datawise_challenges_opportunities}
\end{longtable}

\subsubsection{Additional open topics}

We have discussed the challenges and limitations of designing an automated climate observation framework based on remote sensing and AI methods. These challenges and limitations can be denoted as the `known knowns'. On the other hand, we would like to remind that in the field of climate science there are also `known unknowns', and there are many other `unknown unknowns' waiting to be identified by researchers.

\textbf{Domino effect and self-reinforcing feedback loops}

One of the known unknowns is the domino effect of the threshold events. Steffen et al. \cite{Steffen8252} discussed that there are several temperature thresholds that will trigger climate events, which in turn will start triggering other climate events on another side of the Earth. This is explained as the domino effect in their study. To the best of our knowledge, such self-reinforcing climate triggers are not considered in any climate model. Thus, any predictive model designed so far will be incapable of predicting the conditions after these events are triggered.

\textbf{Reliability of the satellite-based observations}

In their recent study, \cite{Jia2021} showed that satellites have been underestimating the cooling needs of our planet. The reason is that satellite-image-based cloud-cover calculations are mostly based on the lower clouds since they generally work relying on cloud-shadow-detection algorithms. Higher clouds and aerosol-masking effects are not considered in the current Earth radiation needs in order to keep the global average temperatures below 2$^\circ$C \cite{Andreae2005}. In short, the Earth would need more cooling than what has been predicted based on satellite-image-based calculations (where higher clouds and aerosol masking are not taken into account). Considering the fact that virtually all climate-adaptation indicators are extracted from satellite images, we should consider high uncertainties in any AI model which used for future predictions.  

\textbf{Changing behaviour of the same environmental components}

Another uncertainty is associated with the unknown behaviour of the changing environmental components when the temperatures change. When performing our projections for the future, we assume that there is no change in the amount of CO$_2$ that can be stored in oceans and in soil. However, researchers showed that even a 2$^\circ$C increase in soil temperature will produce a significant reduction in the soil storage capacity. The same applies for the CO$_2$ storage capability of the oceans. This implies that even if we managed to reduce our CO$_2$ emissions, there will still be an increased amount of CO$_2$ in the air \cite{Friedlingstein2003}.

\textbf{Changing behaviour of humans and land-use activities}

With increasing heat, more residential, business and other buildings have started using multiple-air conditions. Even in the countries with widely extended cycling habits, in times of heat waves cycling is unsafe because of the danger of a heatstroke. Therefore more people started commuting with air-conditioned cars even for short distances. Furthermore, in the context of the unexpected current pandemic, most of the vaccines require special refrigerators to maintain them in good conditions. These behaviour changes (and many more unexpected habits to come) might have high potential on impact climate change.

\section{Concluding remarks}\label{sec:conclusions}

In this article we have discussed automation possibilities of climate-adaptation observation of cities using remote-sensing data and high-fidelity flow simulations. We have proposed methods where AI can help to perform reliable predictions and also for building finer-scale flow and climate models. We have discussed challenges and limitations of building the proposed models. Our next steps will be focused on creating demonstrations in some test cities within Europe.


\begin{center}
\textit{``If I had more time, I would have written a shorter letter. --- Blaise Pascal"}
\end{center}


   
 
 
 
 
 

\section*{ACKNOWLEDGEMENTS}\label{ACKNOWLEDGEMENTS}
Ricardo Vinuesa acknowledges the financial support of the G\"oran Gustafsson foundation for his contribution.


\bibliography{mybibfile}

\end{document}